# SAR Image Colorization: Converting Single-Polarization to Fully Polarimetric Using Deep Neural Networks

*Qian Song, Student Member, IEEE, Feng Xu, Senior Member, IEEE, and Ya-Qiu Jin, Fellow, IEEE*

*Key Lab for Information Science of Electromagnetic Waves (MoE), Fudan University, Shanghai 200433, China*

*Funding Support: National Key R&D Program of China no. 2017YFB0502700, NSFC no. 61571134*

**Abstract**: A deep neural networks based method is proposed to convert single polarization grayscale SAR image to fully polarimetric. It consists of two components: a feature extractor network to extract hierarchical multi-scale spatial features of grayscale SAR image, followed by a feature translator network to map spatial feature to polarimetric feature with which the polarimetric covariance matrix of each pixel can be reconstructed. Both qualitative and quantitative experiments with real fully polarimetric data are conducted to show the efficacy of the proposed method. The reconstructed full-pol SAR image agrees well with the true full-pol image. Existing PolSAR applications such as model-based decomposition and unsupervised classification can be applied directly to the reconstructed full-pol SAR images. This framework can be easily extended to reconstruction of full-pol data from compact-pol data. The experiment results also show that the proposed method could be potentially used for interference removal on the cross-polarization channel.

**Keywords**: *Polarimetric synthetic aperture radar, SAR image colorization, deep neural network (DNN)*

## 1. Introduction

Synthetic Aperture Radar (SAR), in general, can be classified into four categories in terms of polarimetry capability: single-polarization (single-pol), dual-polarization (dual-pol, e.g. HH/VH, VV/HV), compact-polarization (compact-pol), and full polarimetry (full-pol). Apparently, full-pol system can obtain richer information, but with sacrifice of reduced resolution or swath and increased system complexity. The compact-pol mode is then proposed to make a good compromise among these factors [Souyris et al., 2002]. Only one instead of two transmit/receive cycles is required per along-track sample, thus can achieve a half repetition frequency and data rates or a twice swath width. It hopes to obtain more information than dual-pol by using different polarization basis for transmit and receive, respectively.

In the area of compact-pol SAR image analysis, one approach is to reconstruct a pseudo full-pol data from compact-pol by enforcing additional constraints which are derived from *a priori* assumptions. Thereafter, full-pol processing methods can be applied to the reconstructed data. For example, the reflection symmetry was assumed in [Souyris et al., 2005], which is often observed in forests, but not true for media with anisotropic orientation. Apparently, such *a priori* constraint should be specific to each type of terrain surface. An ideal solution would be to make the constraints adaptive to each pixel. Such pixel-by-pixel adaptive mechanism has never been explored in previous studies. It has to rely on either additional source of information such as landcover map, or features extracted from SAR image itself such as textures. For the latter approach, we are in fact trying to recover the lost polarimetric information by digging into the spatial patterns of grayscale SAR image. Following this approach, one might wonder: is it possible to reconstruct full-pol from single-pol SAR image only?

This paper attempts to reconstruct fully polarimetric SAR image directly from single-pol grayscale SAR image via mapping the extracted spatial feature to the polarimetric feature space. This is a feasible approach because our visual experiences indicate that human can interpret most terrain types by merely looking at the texture. It would be natural for human to reconstruct a colorful image from a gray image, as our visual experiences would automatically relate textures that look like vegetation to green, and textures that look like ocean to blue, etc. Inspired by a stream of studies in computer vision (CV), called 'automatic image colorization', which aims to converting grayscale image into color image without any additional information, this paper proposes 'SAR image colorization', which converts single-pol SAR image to fully polarimetric using deep neural networks (DNN).

Optical image colorization fall into two categories: semi-automatic and fully automatic colorization. Scribble-based method [Levin et al., 2004] is one of the semi-auto colorization methods, which requires user to scribble desired colors in certain regions. Then the colors are propagated to the whole image based on the assumption that the adjacent pixels that have similar intensities should have similar colors. In recent years, DNN was introduced to fully automatic image colorization. Larsson et al. (2016) propose to exploit low-level and semantic

2
Submitted to IEEE TGRS

representations for per-pixel color prediction using DNN. It stacks the features extracted from the VGG16 networks to form a hyper-column for each pixel, and the hyper-column is fed into a three-layer fully connection network to predict color. Simultaneously, Zhang et al. (2016) proposed a similar system which differs in terms of the network architecture and the rebalanced classification loss function.

In this study, we first use convolutional neural networks (CNN) to extract hierarchical multi-scale spatial features from grayscale single-pol SAR image. The multi-scale spatial features are extracted from the intermediate layers of CNN and then stacked up to a hyper-column feature vector. Subsequently, we construct another deep neural network to map the spatial features to the normalized polarimetric feature space so that we can reconstruct a fully polarimetric covariance matrix for each pixel. Fig. 1 depicts the proposed architecture. The first key component is the feature extractor network, which, in this study, is simply consisted of several layers from the pre-trained VGG CNN. The second key component is the feature translator network, which is a deep fully-connected neural network. The networks are trained with patches of full-pol SAR images, i.e. single-pol as input vs full-pol as output, and then tested on other SAR image. The proposed approach is demonstrated with NASA/JPL UAVSAR L-band SAR image. Quantitative metrics are employed to measure the reconstruction error. Visualization analyses show that the feature space is reasonable and feature translation rules can be automatically learned from the data. Note that no additional assumption such as reflection symmetry is made. Although only single-pol-to-full-pol is showcased in this study, the same architecture can be used to convert any form of non-full-pol data to full-pol, including compact-pol and dual-pol.

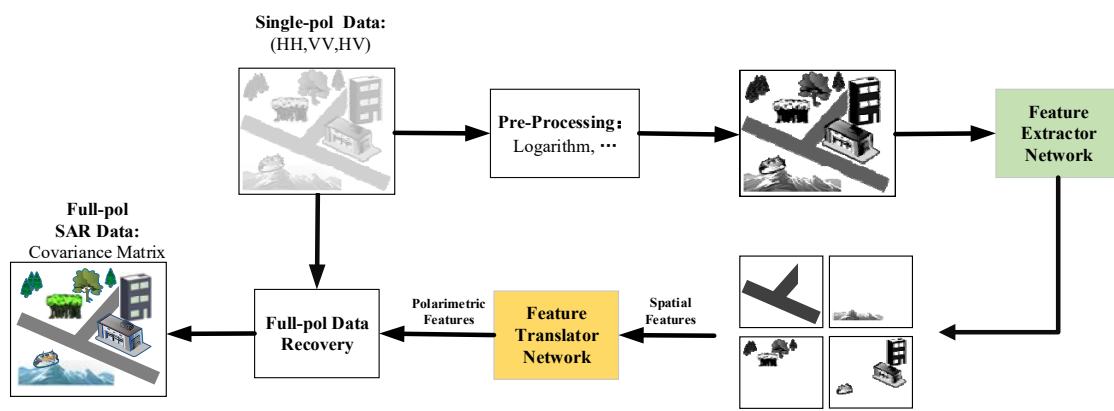

Figure 1 A framework for automatic SAR image colorization.

This paper makes a first attempt to directly map spatial feature in SAR image to polarimetric feature via deep neural networks. The feature mapping mechanism is automatically learned from training data, which requires no *a priori* knowledge or restriction. It establishes a DNN-based general framework to translate different forms of features in SAR image. This framework can be easily extended to conversion of any form of non-full-pol data to full-pol, including compact-pol and dual-pol. Different applications could be developed based on such framework.

The reminder of the paper is organized as follows: Section II first formulates the full-pol reconstruction problem. Section III introduces the technical details of the proposed architecture as well as visualization analyses of the learned networks. Section IV presents testing results along with discussions. Section V concludes the paper.

## 2. Problem Formulation for SAR Colorization

### a. Normalized Polarimetric Feature Space

For fully polarimetric SAR, if the scattering reciprocity $S_{hv} = S_{vh}$ holds, the scattering vector is written as:

$$\vec{k}_L = \begin{bmatrix} S_{hh} & \sqrt{2}S_{hv} & S_{vv} \end{bmatrix}^{\mathrm{T}} \tag{1}$$

where $S_{ij}$ $(i, j = h, v)$ denotes the scattering amplitude function with transmitting polarization $j$ and receiving polarization $i$. Superscript $\mathrm{T}$ denotes transpose. For multilooked complex SAR image, we can use the covariance matrix to solely represent its polarimetric information, that is:

$$\mathbf{C} = \langle \vec{k}_L \cdot \vec{k}_L^{\mathrm{H}} \rangle = \begin{bmatrix} \langle |S_{hh}|^2 \rangle & \sqrt{2}\langle S_{hh}S_{hv}^* \rangle & \langle S_{hh}S_{vv}^* \rangle \\ \sqrt{2}\langle S_{hv}S_{hh}^* \rangle & 2\langle |S_{hv}|^2 \rangle & \sqrt{2}\langle S_{hv}S_{vv}^* \rangle \\ \langle S_{vv}S_{hh}^* \rangle & \sqrt{2}\langle S_{vv}S_{hv}^* \rangle & \langle |S_{vv}|^2 \rangle \end{bmatrix} \tag{2}$$

Where superscript $\mathrm{H}$ denotes conjugate transpose, $\langle \cdot \rangle$ denotes ensemble averaging, which is often taken across different sub-apertures and/or sub-bands. Note that the covariance matrix is Hermitian positive semi-definite matrix.

The trace of covariance matrix $\mathbf{C}$ equals to the Frobenius norm (Span) of the scattering matrix, which physically means the total scattering power $P$:

$$P = \text{Span}(\mathbf{S}) = \text{Tr}(\mathbf{C}) \tag{3}$$

Let's define the ratio of diagonal elements of covariance matrix to $P$ as:

$$\delta_1 = \frac{\langle |S_{hh}|^2 \rangle}{P}, \quad \delta_2 = \frac{2\langle |S_{hv}|^2 \rangle}{P}, \quad \delta_3 = \frac{\langle |S_{vv}|^2 \rangle}{P} \tag{4}$$

Apparently, we have the sum of the ratios equals to 1, i.e. $\delta_1 + \delta_2 + \delta_3 = 1$. In addition, the polarimetric correlation coefficients can be defined as follows:

$$\rho_{13} = \frac{\langle S_{hh} S_{vv}^* \rangle}{\sqrt{\langle |S_{hh}|^2 \rangle \langle |S_{vv}|^2 \rangle}} \tag{5}$$

$$\rho_{23} = \frac{\sqrt{2}\langle S_{hv} S_{vv}^* \rangle}{\sqrt{\langle |S_{hv}|^2 \rangle \langle |S_{vv}|^2 \rangle}} \tag{6}$$

$$\rho_{12} = \frac{\sqrt{2}\langle S_{hh} S_{hv}^* \rangle}{\sqrt{\langle |S_{hh}|^2 \rangle \langle |S_{hv}|^2 \rangle}}. \tag{7}$$

Apparently, the normalized $\mathbf{C}$ can now be expressed in terms of these 6 parameters, where only 5 of them are independent, i.e.

$$\frac{\mathbf{C}}{P} = \begin{bmatrix} \delta_1 & \rho_{12}\sqrt{\delta_1 \delta_2} & \rho_{13}\sqrt{\delta_1 \delta_3} \\ \rho_{12}^*\sqrt{\delta_1 \delta_2} & \delta_2 & \rho_{23}\sqrt{\delta_2 \delta_3} \\ \rho_{13}^*\sqrt{\delta_1 \delta_3} & \rho_{23}^*\sqrt{\delta_2 \delta_3} & \delta_3 \end{bmatrix}. \tag{8}$$

Note that the three parameters $\delta_1$, $\delta_2$, $\delta_3$ are in the range [0,1], while the amplitude of the complex coefficients $\rho_{13}$, $\rho_{23}$, $\rho_{12}$ are in the range [0,1]. Hence, we can define the complex high-dimensional feature space spanned by these six parameters as the normalized polarimetric features, which is independent of the total scattering power. It is a hypersphere with 3 real degrees of freedom and 3 complex degrees of freedom. For convenience, we further define a vectorized version of the normalized polarimetric feature space as

$$\vec{C} = [\delta_1 \quad \delta_2 \quad \delta_3 \quad \rho_{13} \quad \rho_{23} \quad \rho_{12}]^{\text{T}}. \tag{9}$$

### *b. Reconstruction*

The reconstruction problem can be treated as a two-stage problem. First, a feature extractor is used to extract the multi-scale spatial feature from the grayscale single-pol SAR image, i.e.

$$\vec{F}_{i,j} = \mathcal{H}\left(\begin{bmatrix} I_{i-\frac{n}{2},j-\frac{n}{2}} & \cdots & I_{i-\frac{n}{2},j+\frac{n}{2}} \\ \vdots & \ddots & \vdots \\ I_{i+\frac{n}{2},j-\frac{n}{2}} & \cdots & I_{i+\frac{n}{2},j+\frac{n}{2}} \end{bmatrix}\right) \qquad (10)$$

where $I_{i,j}$ denotes the intensity of the $i,j$-th pixel of the grayscale single-pol SAR image, $\vec{F}_{i,j}$ denotes the extracted feature vector of the $i,j$-th pixel and its context. $\mathcal{H}(\cdot)$ denotes the feature extractor network. Apparently, the input to the feature extractor is a patch of image centered at the $i,j$-th pixel with window size of $n \times n$.

Second, the spatial feature vector $\vec{F}_{i,j}$ should be mapped to the polarimetric feature space, i.e.

$$\vec{C}_{i,j} = \mathcal{M}(\vec{F}_{i,j}) \qquad (11)$$

where $\mathcal{M}(\cdot)$ is the feature translator network.

After that, the normalized polarimetric feature is now represented by 3 real ratios + 3 complex coefficients, then the normalized covariance matrix can be reconstructed using Eq. (8). For example, for a VV-pol grayscale SAR image, the total power can be recovered as:

$$P = \frac{\langle |S_{vv}|^2 \rangle}{\delta_3} \qquad (12)$$

and subsequently the covariance matrix can be constructed.

The purpose of SAR image colorization is to fit the hierarchical networks of $\mathcal{H}(\cdot)$ and $\mathcal{M}(\cdot)$ with polarimetric data and then test it on single-pol grayscale data.

### c. Correction for Positive Semi-definiteness

The scattering vector can be modeled as having complex Gaussian distribution, and the ensemble averaged covariance matrix in Eq. (2) is, by definition, positive semi-definite. The reconstruction formulation in Eq. (8) guarantees the reconstructed covariance matrix to be a Hermitian matrix, i.e. $\mathbf{C} = \mathbf{C}^{\mathrm{H}}$. However, the reconstructed $\mathbf{C}$ may not be positive semi-definite as there is no such constraint being applied to the reconstructed six parameters. Many algorithms used to process fully polarimetric data such as eigen-analysis decomposition rely on such property of the covariance or coherency matrix. Thus, this paper proposes an algorithm to correct it for the positive semi-definiteness.

According to the Sylvester's criterion [C. D. Meyer, 2000], the Hermitian matrix $\mathbf{C}$ is positive semi-definite if all of its leading principal minors are positive or equal to zero, i.e. the determinants of $\mathbf{C}$ and its upper-left 1x1 and 2x2 corner submatrices are positive or equal to zero. Hence, the positive semi-definiteness constraints of $\mathbf{C}$ in Eq. (8) can be written as

$$\det([\delta_1]) = \delta_1 \geq 0$$

$$\det\begin{bmatrix} \delta_1 & \rho_{12}\sqrt{\delta_1\delta_2} \\ \rho_{12}^*\sqrt{\delta_1\delta_2} & \delta_2 \end{bmatrix} = 1 - |\rho_{12}|^2 \geq 0 \quad (13)$$

$$\det\left(\frac{\mathbf{C}}{P}\right) = 1 + 2\mathrm{Re}(\rho_{12}\rho_{23}\rho_{13}^*) - |\rho_{13}|^2 - |\rho_{23}|^2 - |\rho_{12}|^2 \geq 0$$

where $\det$ denotes taking the determinant and $\mathrm{Re}$ denotes taking the real part. Apparently, the first two constraints are already satisfied by definition. The last inequality is the only constraint to make $\mathbf{C}$ positive semi-definiteness. It can be found that it only involves with the 3 complex coefficients.

This paper proposes an algorithm to correct $\rho_{12}, \rho_{23}$ so that the inequality in Eq. (13) can be satisfied. Note that $\rho_{13}$ the correlation between co-polarization is believed to be more informative than the correlations with cross-polarization $\rho_{12}, \rho_{23}$. Hence, this algorithm only correct $\rho_{12}, \rho_{23}$ but not $\rho_{13}$.

For simplicity, let's use $r_{ij}$ and $\varphi_{ij}$ to denote the amplitude and phase of $\rho_{ij} = r_{ij}\exp(j\varphi_{ij})$, $r_{ij} = |\rho_{ij}|$. With certain algebraic simplification, the last inequality of Eq. (13) can be rewritten as

$$R = \frac{r_{13}^2 + r_{23}^2 + r_{12}^2 - 1}{2r_{12}r_{13}r_{23}} \leq \cos(\varphi_{12} + \varphi_{23} - \varphi_{13}) \leq 1 \quad (14)$$

We can take a two-step correction procedure to make sure that this inequality holds:

a) In the first step, we correct $r_{12}, r_{23}$, if necessary, to guarantee $R \leq 1$. It is found that we can always achieve it by reducing the amplitudes $r_{12}, r_{23}$ by a factor of $\eta \in [0,1]$, i.e. $r_{23} \Leftarrow \eta r_{23}$, $r_{12} \Leftarrow \eta r_{12}$. The factor $\eta$ can be easily solved for by enforcing $R = 1$ as

$$\eta = \begin{cases} \sqrt{\dfrac{1 - r_{13}^2}{r_{23}^2 + r_{12}^2 - 2r_{13}r_{23}r_{12}}} & R > 1 \\ 1 & R \leq 1 \end{cases} \qquad (15)$$

Note that the condition $\eta < 1$ holds if $R > 1$.

b) In the second step, since $R \leq 1$ is true, we can always correct $\varphi_{12}, \varphi_{23}$ to guarantee the inequality of Eq. (14), i.e. $\varphi_{12} \Leftarrow \varphi_{12} + \Delta\varphi/2$, $\varphi_{23} \Leftarrow \varphi_{23} + \Delta\varphi/2$. Likewise, the adjustment to the phase angles $\Delta\varphi$ can be solved for by enforcing the equality, i.e.

$$\Delta\varphi = \begin{cases} 0 & if \ \cos(\varphi_{12} + \varphi_{23} - \varphi_{13}) \geq R \\ \operatorname{acos} R - \varphi_{23} - \varphi_{12} + \varphi_{13} & else \end{cases} \qquad (16)$$

The algorithm of correction for positive semi-definiteness is summarized in Fig. 2.

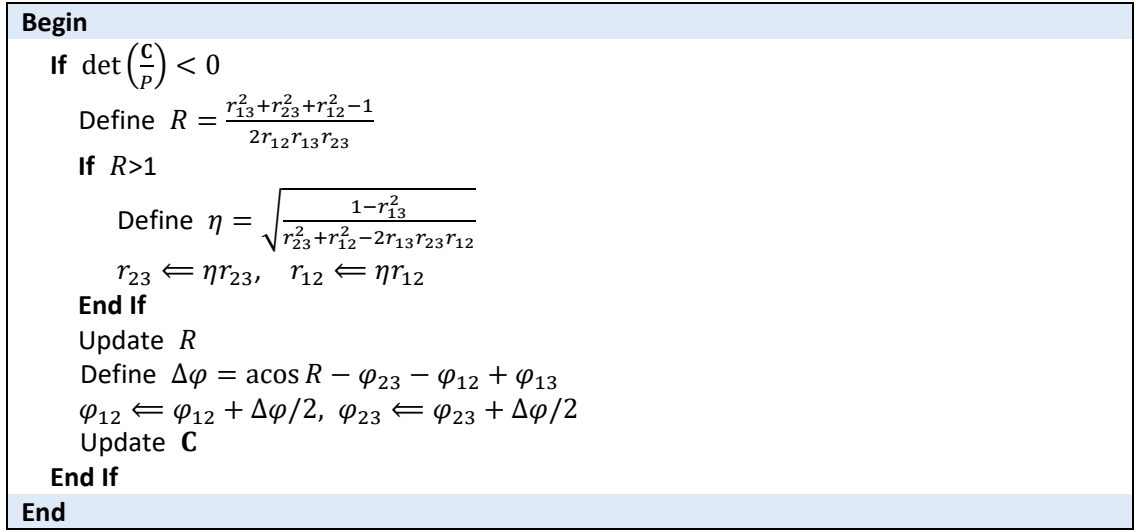

**Begin**
  **If** $\det\left(\dfrac{\mathbf{C}}{P}\right) < 0$
    Define $R = \dfrac{r_{13}^2 + r_{23}^2 + r_{12}^2 - 1}{2r_{12}r_{13}r_{23}}$
    **If** $R > 1$
      Define $\eta = \sqrt{\dfrac{1 - r_{13}^2}{r_{23}^2 + r_{12}^2 - 2r_{13}r_{23}r_{12}}}$
      $r_{23} \Leftarrow \eta r_{23}, \quad r_{12} \Leftarrow \eta r_{12}$
    **End If**
    Update $R$
    Define $\Delta\varphi = \operatorname{acos} R - \varphi_{23} - \varphi_{12} + \varphi_{13}$
    $\varphi_{12} \Leftarrow \varphi_{12} + \Delta\varphi/2, \ \varphi_{23} \Leftarrow \varphi_{23} + \Delta\varphi/2$
    Update $\mathbf{C}$
  **End If**
**End**

Figure 2 Algorithm of correction for positive semi-definiteness of reconstructed covariance matrix.

## 3. Deep Neural Networks for SAR Colorization

### a. Network Architecture

We propose to address the SAR colorization problem with deep neural networks. Recently, DNNs have been widely applied in SAR image analyses [e.g. Chen et al. 2016; Wagner 2016; Zhou et al. 2016; Jiao and Liu, 2016; Zhang et al. 2017]. The novel network architecture proposed in this paper is shown in Fig. 3.

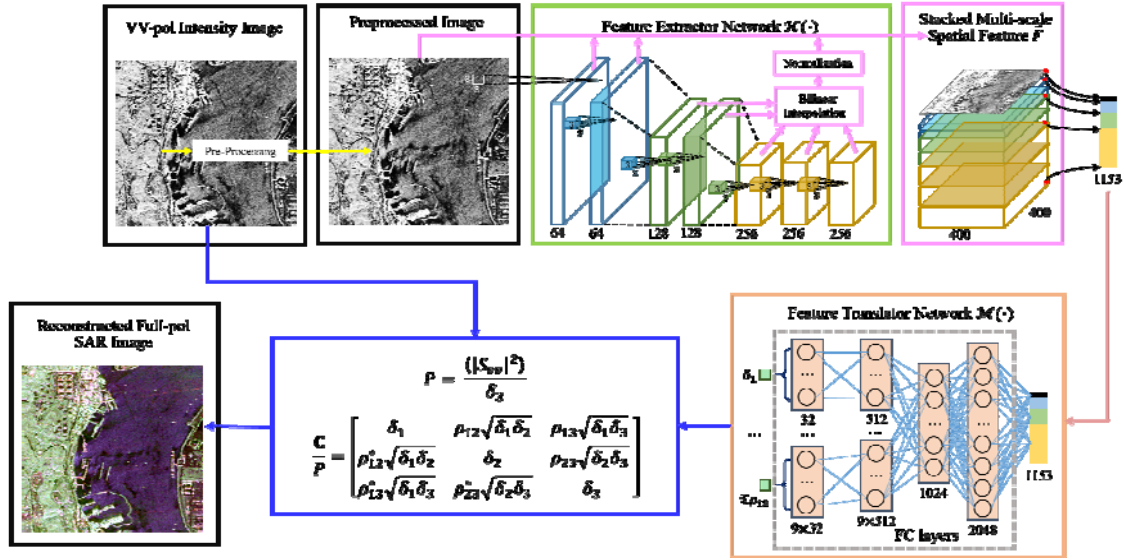

Figure 3 Network Architecture of Full-pol SAR Data Reconstruction from Single-pol Data.

Feature extractor network $\mathcal{H}(\cdot)$ is composed of seven cascaded convolutional layers which extract multiscale spatial features from spatial dimensions to feature dimension layer by layer. Thus, the higher layers represent higher-level/larger-scale features. In this study, we use the first seven convolutional layers of the VGG16 network [Simonyan et al., 2015], which is pre-trained using the ImageNet database [Russakovsky et al., 2015]. Different from optical image colorizations, pixel values in SAR image vary in a large range, so a pre-processing step is crucial to make SAR image suitable to fed into VGG16. In this paper, we firstly calculate logarithm of the intensity image, then the values are linearly projected into [0,1] from [-25, 0]. Since VGG16 is designed for RGB three-channel color images. The weights of VGG16's input layer are averaged across three channels to allow input data of only one channel. Such a pre-trained network can reduce the number of parameters significantly so as to prevent the network from overfitting, while effectively extract the semantic and geometric features. Detailed configuration information about the feature extractor network is given in Table I. It mainly consists of repeated patterns of convolution, ReLU and max-pooling, where ReLU denotes the rectified linear unit nonlinearity.

**Table I Configuration of Feature Extractor Network**

| name | kernel size | stride |
|---|---|---|
| Conv1-1 | 64@ $3 \times 3$ | 1 |
| ReLU1-1 | -- | -- |
| Conv1-2 | 64@ $3 \times 3$ | 1 |
| ReLU1-2 | -- | -- |
| Pool1 | $2 \times 2$ | 2 |
| Conv2-1 | 128@ $3 \times 3$ | 1 |
| ReLU2-1 | -- | -- |
| Conv2-2 | 128@ $3 \times 3$ | 1 |
| ReLU2-2 | -- | -- |
| Pool2 | $2 \times 2$ | 2 |
| Conv3-1 | 256@ $3 \times 3$ | 1 |
| ReLU3-1 | -- | -- |
| Conv3-2 | 256@ $3 \times 3$ | 1 |
| ReLU3-2 | -- | -- |
| Conv3-3 | 256@ $3 \times 3$ | 1 |
| ReLU3-3 | -- | -- |

As the colorization task is conducted on a pixel-by-pixel basis, we have to use multi-scale hierarchical features, which are extracted from all levels. Therefore, the outputs of high-level convolutional layers are first interpolated back to the input image size using bilinear rule. As SAR image in general differs greatly from optical images in ImageNet database, the scale of extracted feature values may vary greatly. This justifies a normalization operation applied to each feature layer across all training images. This is a conventional measure taken by DNN-based optical image colorizations as well [e.g. Liu et al., 2015]. It is defined as

$$\vec{F}_{i,j}(k) \Leftarrow \frac{\vec{F}_{i,j}(k) - \mu_k}{\sigma_k} \qquad (17)$$

$\vec{F}_{i,j}(k)$ is the $k$-th layer feature vector of $i,j$-th pixel, and $\mu_k$ and $\sigma_k$ denotes the estimated mean and standard deviation of extracted features of the $k$-th layer, respectively. After normalization, features from different layers could contribute equally to the next stage network.

Features of the corresponding pixel from all layers including the input image itself are then concatenated together to form a hyper-column descriptor which is used as the desired final feature vector $\vec{F}_{i,j}$.

On the second stage, the hyper-column feature vector is fed into a five-layer fully-connected neural network which plays as the role of feature translator $\mathcal{M}(\cdot)$. Its layer size is 1153, 2048, 1024, 512 and 32 respectively, as shown in Table II. Each layer is followed by a ReLU nonlinearity. The output layer of feature translator network is critical. Conventionally, regression-type of output neuron should be used to predict a continuous value. However, we found that it is better to use classification-type output to predict discretize levels of a continuous value. Hence, softmax output layer is employed. This is explained in next subsection.

**Table II    Configuration of Feature Translator Network**

| FC1: 1153 ⇨ 2048 | | | |
|---|---|---|---|
| ReLU1 | | | |
| FC2: 2048 ⇨ 1024 | | | |
| ReLU2 | | | |
| FC3-1-1: 1024 ⇨ 512 | FC3-2-1: 1024 ⇨ 512 | ⋯ | FC3-9-1: 1024 ⇨ 512 |
| ReLU3-1 | ReLU3-2 | ⋯ | ReLU3-9 |
| FC3-1-2: 512 ⇨ 32 | FC3-2-2: 512 ⇨ 32 | ⋯ | FC3-9-2: 512 ⇨ 32 |
| Softmax1 | Softmax2 | ⋯ | Softmax9 |

### b. Non-uniform Quantization

The output layer of feature translator $\mathcal{M}(\cdot)$ consists of 9 independent groups of neurons, each of which predicts one of the 9 unknown polarimetric parameters, respectively. DNN is known to be good at classification tasks. Hence, each group of output neurons is in fact a classification-type output (softmax) used to predict discrete levels of the corresponding parameter. Therefore, the output values have to be first quantized into K bins, and reconstruction can be seen as a prediction of probability distribution over the K bins. This matches with the principle of softmax output. According to our analysis, $K = 32$ bins is sufficient where the error caused by quantization is negligible. For the 3 real parameters, the quantization is done within the [0,1] range, while it is within the [-1,1] range for the real and imaginary part of complex parameters. Note that the amplitudes of 3 complex parameters is limited to [0,1].

In this study, non-uniform quantization is employed to balance the training samples for classification task. Histogram equalization strategy is used for non-uniform quantization. Figure 3 shows an example of non-uniform quantization of parameter $\delta_3$ in one of our experiments.

Results shown later suggest that the 32-level quantization is sufficient to preserve the accuracy of polarimetric information.

With the softmax output probability distribution over the discretized bins, a continuous value of the parameter can be recovered simply by using the conditional mean. However, in this paper the mode of output distribution is used as the final reconstructed parameter, which gives rise to a better visual effect of the reconstructed image.

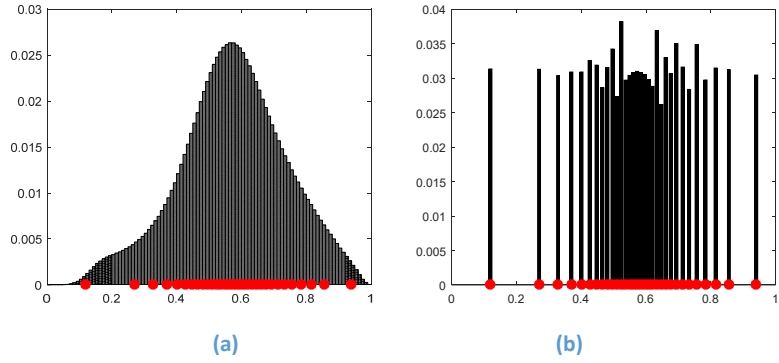

(a)          (b)

Figure 4 Non-uniform quantization of parameter $\delta_3$: (a) histogram of uniform bins; (b) the recalculated histogram of 32 non-uniform bins, where the red points denote the centers of each bin.

### c. Loss Function and Training

The softmax output layer of the feature translator network $\mathcal{M}(\cdot)$ is defined as:

$$p_i^{(j)}(k) = \frac{\exp\left(\mathcal{M}_i^{(j)}(k)\right)}{\sum_{k=1}^{32}\exp\left(\mathcal{M}_i^{(j)}(k)\right)} \tag{18}$$

where $p_i^{(j)}(k)$ denotes the probability of $j$-th parameter ($j = 1,2,\ldots,9$) of $i$-th pixel being in the $k$-th bin (i.e. $k$-th center quantization value). Cross-entropy loss function should be used for training. The overall loss function is defined as:

$$L(y;\boldsymbol{\theta}) = -\frac{1}{9BK}\sum_{i=1}^{B}\sum_{j=1}^{9}\sum_{k=1}^{K}\delta\left(y_i^{(j)} = k\right)\ln p_i^{(j)}(k), \tag{19}$$

where $\boldsymbol{\theta}$ denotes the parameters of the network. $B$ is the total number of pixels in one mini-batch ($B = 2000$ in this paper). $y_i^{(j)}$ denotes the true bin index of $j$-th parameter of $i$-th pixel. $\delta(\cdot)$ is indicator function, and equals to 1 if $y_i^{(j)} = k$, and 0 otherwise. Multiple loss functions of all pixels and all parameters are equally back-propagated through the layers of DNN:

$$\frac{\partial L(\boldsymbol{\theta})}{\partial \boldsymbol{\theta}} = -\frac{1}{9BK}\sum_{i=1}^{B}\sum_{j=1}^{9}\sum_{k=1}^{K}\left[1 - \delta\left(y_i^{(j)} = k\right)p_i^{(j)}(k)\right]\frac{\partial \mathcal{M}_i^{(j)}(k)}{\partial \boldsymbol{\theta}} \qquad (20)$$

As for the training algorithm, stochastic gradient descent (Adam) is used with momentum parameters $\beta_1 = 0.9$ and $\beta_2 = 0.999$ and regularization parameter $\epsilon = 10^{-6}$, and learning rate 0.0001. Extracting hyper-column feature vector for an entire image (1000×1000 would require a memory of 3.93GB) is costly. SAR image is handled as 400x400 patches in both training and test. For each patch, pixels are grouped into mini-batches of 2000 pixels and these mini-batches are randomly sorted to train the network.

As shown in Fig. 5, a full-pol SAR image taken over San Diego, CA by NASA/JPL UAVSAR [NASA, 2012] is used as a case study. The regions including various types of terrain as highlighted by yellow boxes are used as the training data. Test results are given in the next section.

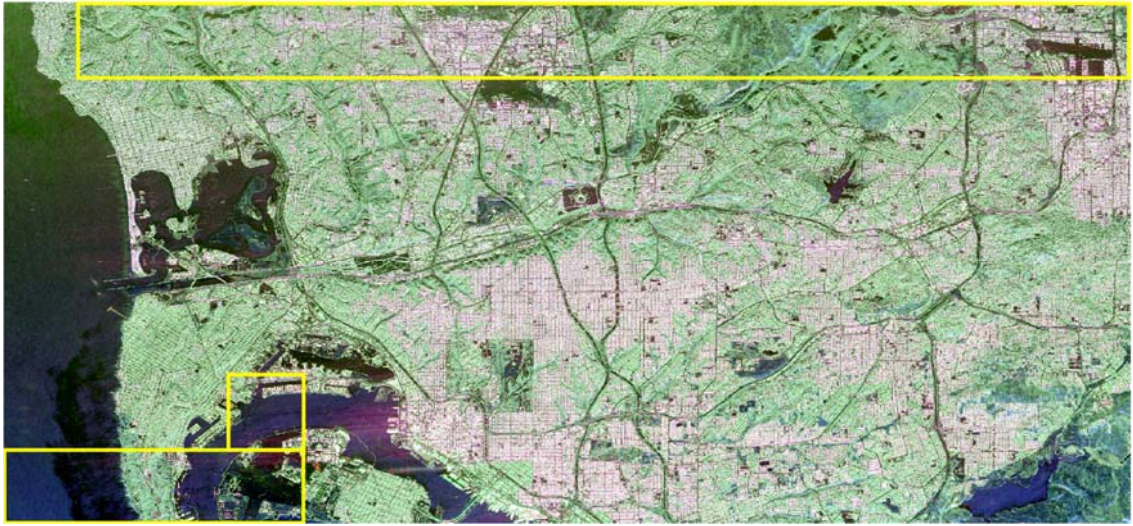

Figure 5    The Pauli-color-coded UAVSAR image of San Diego.

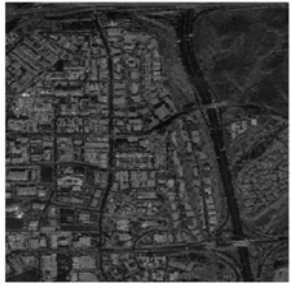

(a) VV-pol Grayscale SAR image of One Patch of Training Data

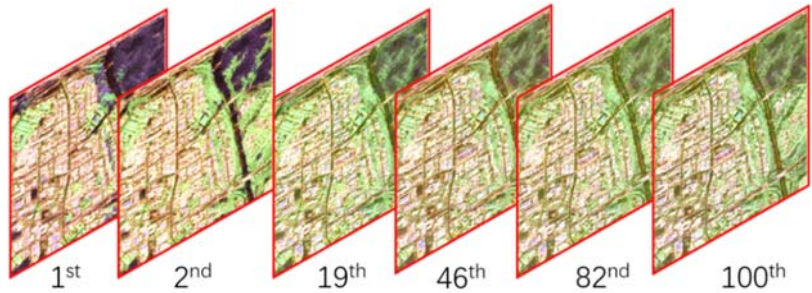

(b) Visualization of Reconstructed Full-pol Data During Training

Figure 6 Visualization of the Training Process

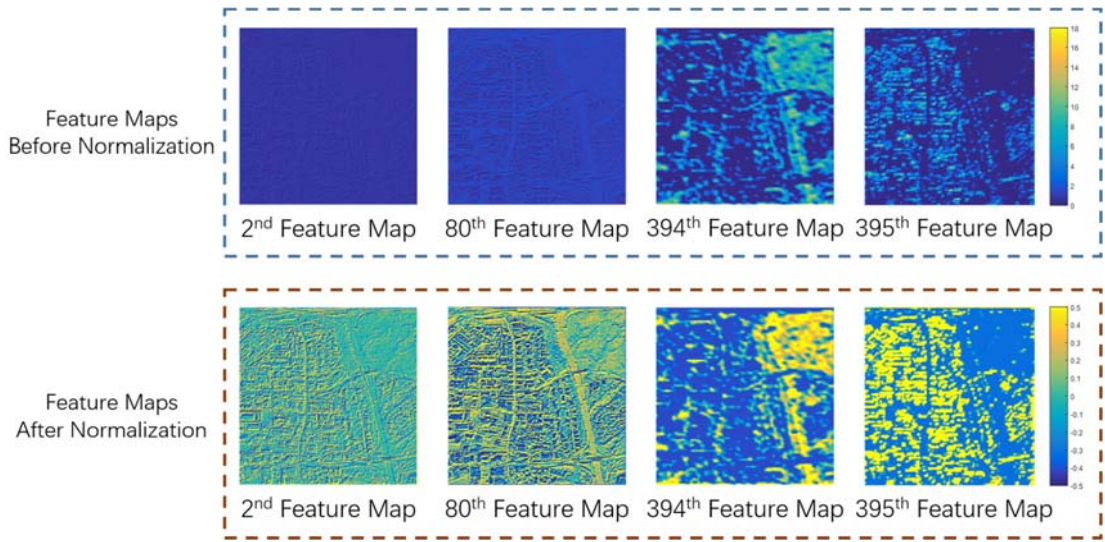

Figure 7 Examples of Normalization Applied to Feature Maps

Fig. 6 (a) is one patch of training data shown in Fig. 5. Fig. 6 (b) shows the color-coded fitted full-pol data during training. It's interesting that the network first learns the segmentation information, and then learn to colorize the SAR image more and more precisely. Fig. 7 presents

Submitted to IEEE TGRS

an example demonstrating the efficacy of feature normalization. The feature maps of input single-pol image differ dramatically. The 2$^{nd}$ and 80$^{th}$ feature maps are mostly within 0 to 1, and the 395$^{th}$ feature map ranges from 0 to 6, but the 394$^{th}$ feature map ranges from 0 to 18. However, after normalization operation, all of them show distinct texture information. It can be found that the 395$^{th}$ feature map represents large-scale feature, while the 80$^{th}$ feature maps reflects small-scale feature.

### d. Visualization of the Learned Networks

The learned networks can extract hierarchical representations of spatial and structural information of the input data. This subsection aims to reveal the networks' mechanism via visualizing its internal states at different stages. Three typical terrain types, namely, urban built-up area, vegetation area and sea surface, are selected as the subjects for visualization.

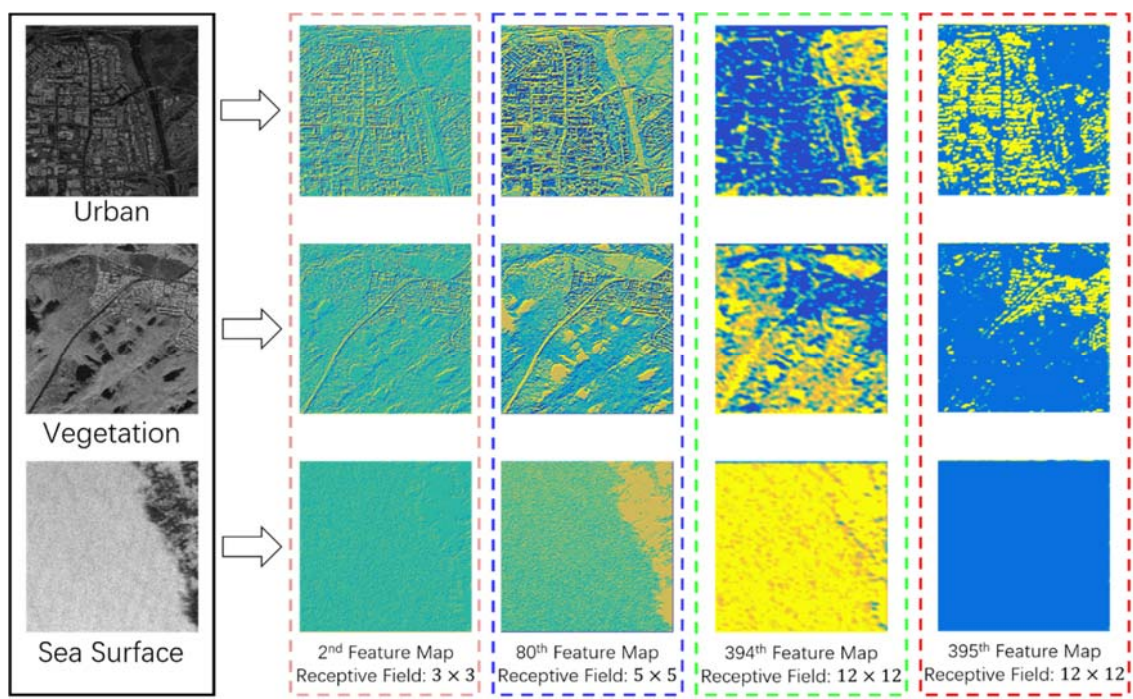

Figure 8 Feature Maps of Three Types of Terrains Extracted by $\mathcal{H}$ Network

Fig. 8 visualizes 4 extracted multi-scale feature maps of the selected 3 typical terrain types. It can be found that feature maps extracted distinct texture information: the 2$^{nd}$ feature map shows the edges of the buildings and roads; the 80$^{th}$ feature map mainly reflects homogeneous surface; the 394th feature map reveals random scattering mechanism; and the 395th feature map correlates to the roads located between buildings.

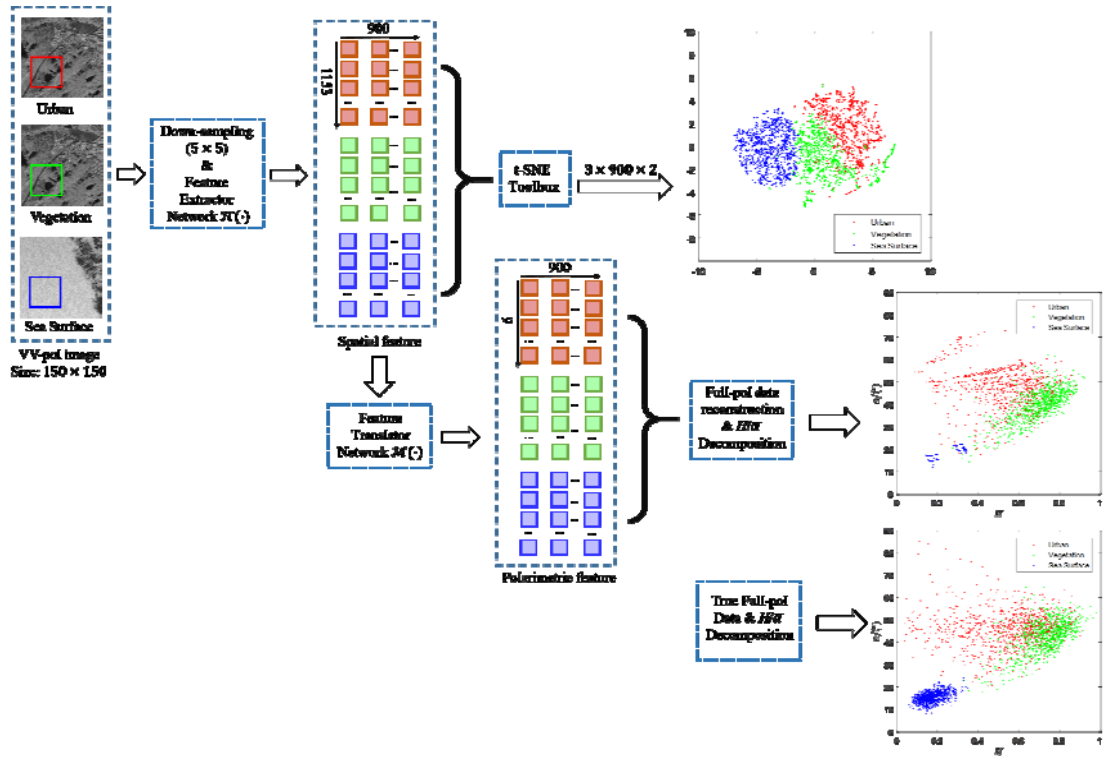

Figure 9 Visualization of Spatial and Polarimetric Features

Fig. 9 shows the visualization of spatial and polarimetric features of the selected 3 regions. The t-Distributed Stochastic Neighbor Embedding (t-SNE) [L.J.P. van der Maaten, 2008], a toolbox for dimensional reduction, is used to map the high-dimensional spatial feature vector onto a 2D plane. Note that t-SNE is an unsupervised technique, and no label information is required. It can be found that 3 terrain types are well separated from each other in t-SNE-mapped 2-D feature space. It demonstrates the efficacy of the feature extractor network and it shows that the extracted features can faithfully represent the intrinsic properties of terrain surfaces.

In order to further show the efficacy of the feature translator network, the well-known $H/\bar{\alpha}$ target decomposition method [Cloude et al., 1997] is employed to visualize the translated polarimetric features of the 3 regions. Entropy $H \in [0,1]$ indicates the randomness of media, and $\bar{\alpha} \in [0,90]$ relates to the scattering mechanisms. As illustrated in Fig . 9, the feature translator network successfully mapped the spatial feature into the correct regions of each terrain surfaces. For example, the sea surface is mapped onto the lower-left corner of single-bounce scattering with low randomness; the vegetation is mapped onto the mid-right region of dipole scattering with high randomness; the urban area is mapped onto the upper region of double-bounce scattering. The true $H/\bar{\alpha}$ map is also plotted side-by-side for comparison, which is calculated directly from the original full-pol data. Interestingly, the distributions of 3 regions appears to be very similar, except that the reconstructed data tends to be more concentrated.

## 4. Results and Analyses

In our experiments, airborne L-band high-resolution polarimetric SAR images obtained by UAVSAR of NASA/JPL are used. The size of training data and test data and other detailed information are listed in Table III. In this paper, three experiments according to three different types of data are conducted. The Test1 and Test2 data are obtained at the same time as the training data. However, the terrain type of Test2 differs from that of training data. While, Test3 is obtained over different area at different time. And the format of Test3 (MLC) is also different from others (GRD). In this section, the qualitative evaluation of the experiments results are firstly presented, followed by quantitative assessment of the method, and two application cases are described at last.

Table III   Training Data and Test Data

|  | Region | Frequency | Data Size | Date | Data Format |
|---|---|---|---|---|---|
| Training | San Diego | L | 400×400×19 | Nov. 9th   2012 | GRD |
| Test1 | San Diego | L | 2800×6000 | Nov. 9th   2012 | GRD |
| Test2 | San Diego | L | 2800×4600 | Nov. 9th   2012 | GRD |
| Test3 | New Orleans | L | 700×700 | Jul. 2nd   2012 | MLC |

### a. Testing Results

In first experiment, the trained networks are applied to convert the VV-pol grayscale SAR image of San Diego, CA obtained at November 9th 2011. Fig. 10(a) shows the VV-pol grayscale image used as input and Fig. 10(b) shows the Pauli-color-coded image of reconstructed full-pol data. The reconstructed image appears to be very close to the true full-pol data (shown in Fig. 5). Note that the top left area in the true image (Fig. 5) appears to be contaminated by sidelobe or radio frequency interference, in particular in the cross-pol HV channel. Nevertheless, the reconstructed full-pol image appears to be contamination-free. This can be explained by the fact that the selected training patches of sea surface are clean ones (shown in the bottom-left of Fig.5). It suggests a potential application of the proposed scheme for interference removal in the low-Signal-to-Noise Ratio (SNR) cross-pol channel by leveraging information of high-SNR co-pol channel. In addition, we also notice that the proposed method is powerful in discriminating texture and context of terrain surfaces. For example, as shown in Fig. 10(a), area A (vegetation) and area B (water) have very similar intensities in VV-channel, and it is difficult for one to interpret them. Nevertheless, the proposed method is able to capture the subtle difference in textures and correctly colorize A to vegetation and B to water surface in Fig. 10(b).

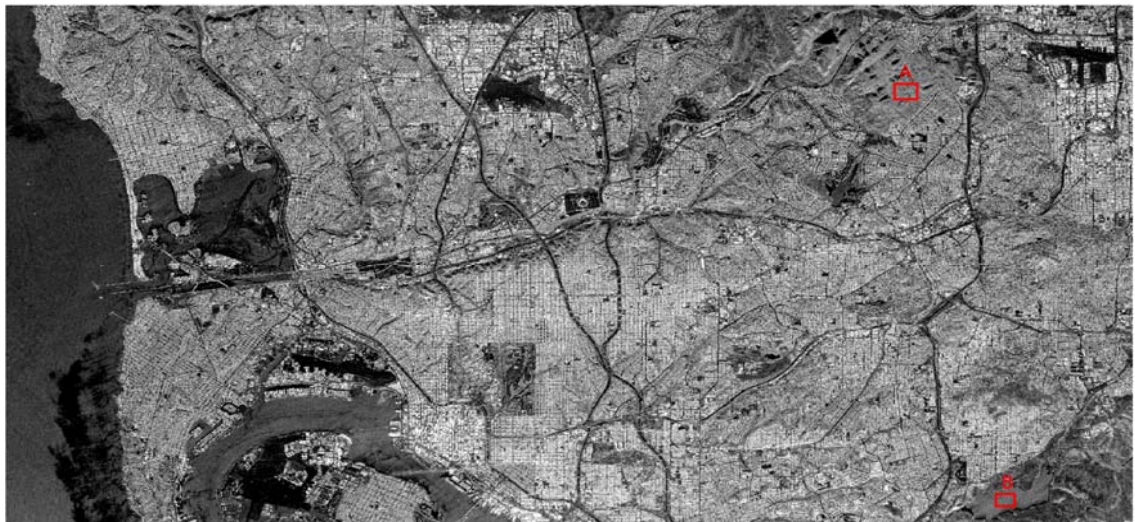

(a) Input VV-pol SAR image

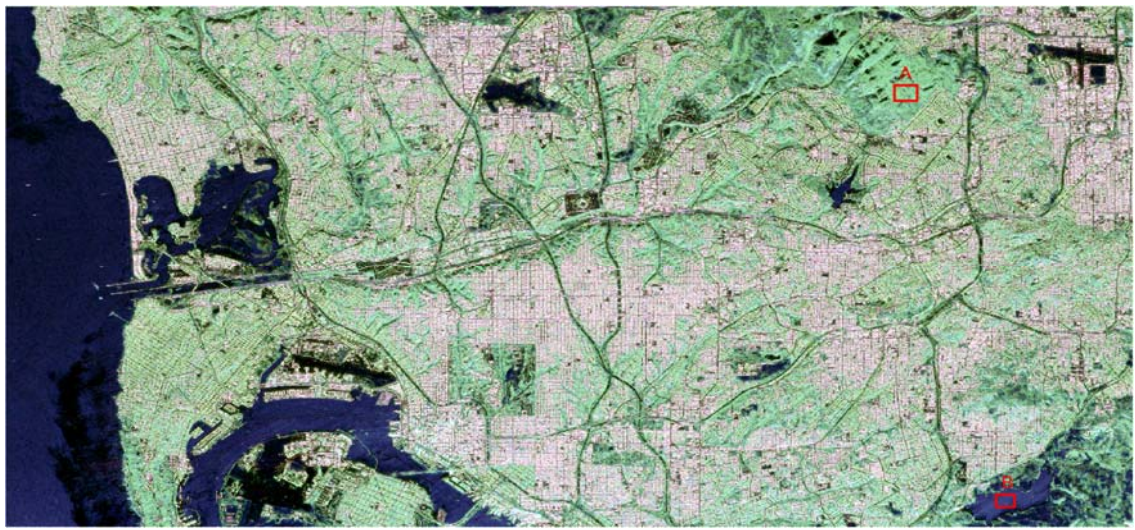

(b) Output full-pol image

Figure 10. Demonstration of SAR image colorization of the UAVSAR image of Test1.

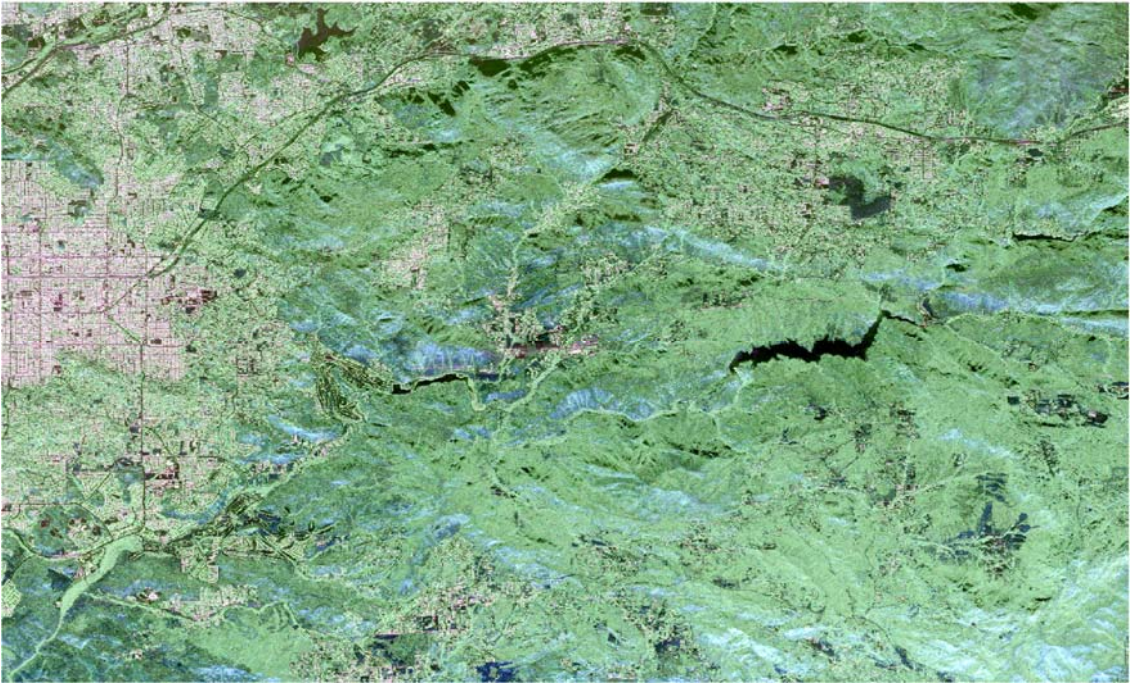

(a) The true full-pol SAR image of Test2 dataset

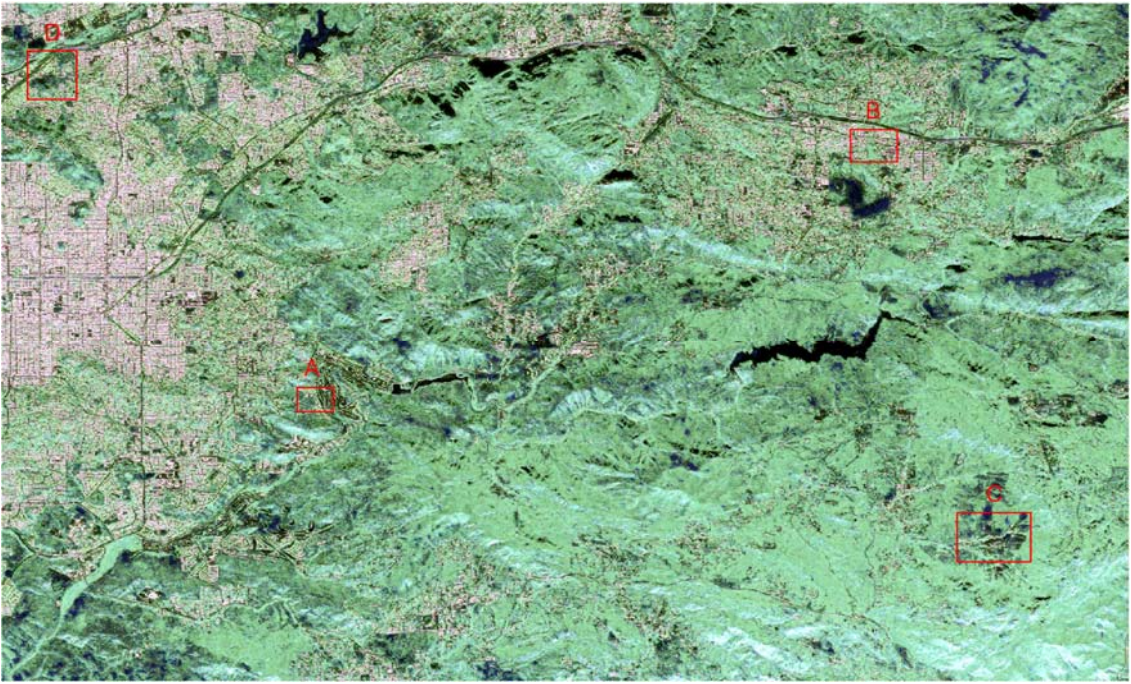

(b) The reconstructed full-pol image of Test2 using VV-pol as input

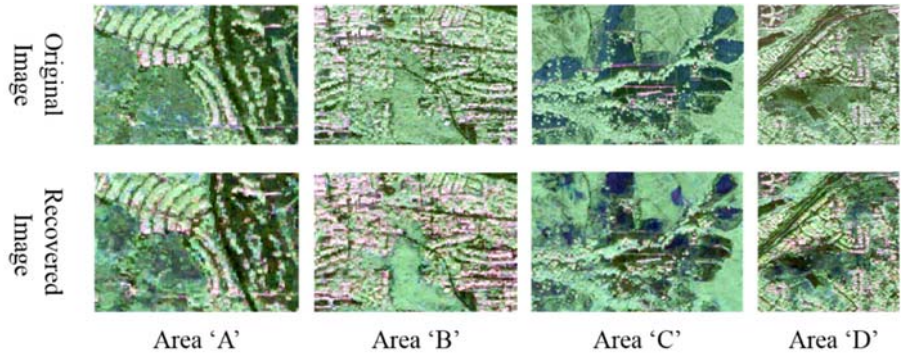

(c) Comparison of true and reconstructed full-pol image of selected areas

Figure 11. Demonstration of SAR colorization of UAVSAR image Test2.

The second case is region mixed of urban, forest and lakes. Fig. 11 (a)(b) show the true full-pol, and the recovered full-pol images using only VV-channel, respectively. Areas marked by red boxes in Fig. 11 (b), are mixed areas containing various terrain objects e.g. buildings, grassland and roads, and Fig. 11 (c) compares the true and recovered full-pol images of these areas. Except that some discrepancies can be observed in vegetation area 'C', most terrain objects including small targets such as narrow roads are well reconstructed. Note that no loss of resolution is observed at all, which demonstrates the efficacy of the multi-scale feature extractor network. This case also shows that the proposed method has successfully captured underlying physical scattering mechanisms. Such approach is potentially useful for target classification or image segmentation [Chen et al. 2016; Zhou et al. 2016].

To further test the robustness of the proposed approach, the Test3 data obtained over New Orleans at 2$^{nd}$ July 2012 is used for testing. This area consists of river, park and buildings. The original true and recovered Pauli-color-coded images are compared in Fig. 12. The reconstructed image agrees well with the true image. The main discrepancies lie in low-intensity areas. As previously discussed, these low-SNR areas are easily contaminated. Water surface in the true full-pol image in Fig. 12 (a) appears to be severely contaminated by sidelobes of nearby vegetation and building scatterers, and thus appears 'greenish' and 'reddish'. However, the reconstructed water surface appears in the reasonable blue color. In some sense, the reconstructed image is of higher quality than the actual measured full-pol one. Note that the 'red' double-scattering of ships in the river are not correctly reconstructed because of the lack of sufficient training samples of such kind.

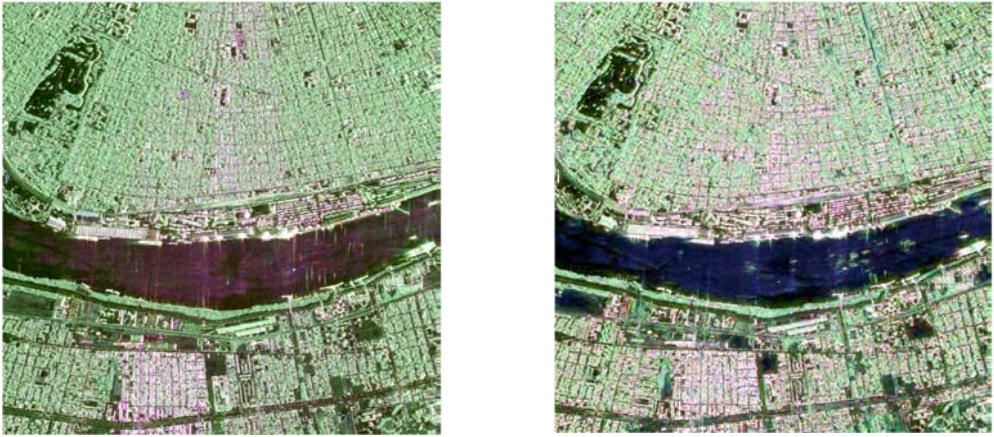

(a) The true full-pol SAR image of Test3     (b) The reconstructed full-pol image of Test3

Figure 12 Demonstration of SAR colorization of UAVSAR image Test3.

## b. *Error Analysis*

In our experiments, there are two sources of errors $\epsilon$: quantization error $\epsilon_q$ and reconstruction error $\epsilon_c$. Quantization error is caused by quantizing the feature values. The comparison of quantization error between uniform $\epsilon_q^*$ and non-uniform $\epsilon_q$ quantizing is also discussed. Reconstruction error comes from the underestimation and/or overestimation of polarimetric features by the networks. Three indicators are used in this paper to quantitatively evaluate similarity between the reconstructed image and the true one:

- Mean absolute error (MAE):

$$\text{MAE}(A, B) = \frac{1}{MN} \sum_{i=1}^{M} \sum_{j=1}^{N} |A_{ij} - B_{ij}| \qquad (21)$$

where $M, N$ are the size of matrix *A* and *B*.

- Coherency Index (COI):

$$\text{COI}(A, B) = \frac{\sum_{i=1}^{M} \sum_{j=1}^{N} A_{ij} B_{ij}^*}{\sqrt{\sum_{i=1}^{M} \sum_{j=1}^{N} A_{ij} A_{ij}^* \times \sum_{i=1}^{M} \sum_{j=1}^{N} B_{ij} B_{ij}^*}} \qquad (22)$$

where superscript $*$ denotes conjugate. COI varies from 0 to 1, indicating the consistency of two complex image.

- Bartlett Distance:

$$d_b(A, B) = 2\ln\frac{\det\left(\frac{A+B}{2}\right)}{\sqrt{\det(A)\det(B)}} \quad (23)$$

Bartlett distance [Kersten et al., 2005] describes the similarity of two covariance matrices of SAR data. The distance would be equal to 0 if and only if $A = B$, or $d_b > 0$.

Table IV shows the results of MAE and COI indicators of Test1. It suggests that the total error mainly comes from reconstruction. The quantization error ranges from 0.002 to 0.016, and non-uniform quantization error is smaller than uniform quantization error for most parameters. For the reconstruction error, the MAEs of three diagonal real parameters are around 0.1, while that of the complex parameters are relatively high. In terms of COI, the reconstructed co-pol channels i.e. $C_{11}, C_{13}$ appears to be accurate, while cross-pol-related channels, in particular $C_{22}, C_{12}$, are less accurate. However, these terms have limited impact in terms of final applications of polarimetric SAR image such as target decomposition or terrain classification.

**Table IV Quantitative Error Analysis**

|  | MAE | | | COI | |
|---|---|---|---|---|---|
|  | $\epsilon_q^*$ | $\epsilon_q$ | $\epsilon$ | $\epsilon$ | |
| $\delta_1$ | 0.0078 | 0.0087 | 0.1345 | $C_{11}$ | 0.7424 |
| $\delta_2$ | 0.0076 | 0.0023 | 0.0414 | $C_{22}$ | 0.3091 |
| $\delta_3$ | 0.0078 | 0.0095 | 0.1194 | $C_{33}$ | 1.0 |
| $\mathbf{Re}(\rho_{13})$ | 0.0156 | 0.0164 | 0.3256 | $C_{13}$ | 0.5500 |
| $\mathbf{Im}(\rho_{13})$ | 0.0156 | 0.0135 | 0.2867 | | |
| $\mathbf{Re}(\rho_{23})$ | 0.0156 | 0.0134 | 0.3180 | $C_{23}$ | 0.4919 |
| $\mathbf{Im}(\rho_{23})$ | 0.0156 | 0.0132 | 0.2906 | | |
| $\mathbf{Re}(\rho_{12})$ | 0.0156 | 0.0141 | 0.3124 | $C_{12}$ | 0.2927 |
| $\mathbf{Im}(\rho_{12})$ | 0.0156 | 0.0130 | 0.3085 | | |

Fig. 13 plots histogram of the pixel-to-pixel Bartlett Distance between the reconstructed and true covariance matrix. It shows that the majority of distances is smaller than 2. The reconstructed data is consistent with true data for the most part, especially for homogeneous terrain such as sea and lakes. Significant discrepancy mainly locates in sea surface along the coast, which is contaminated by the volume and double-bounce scatterers.

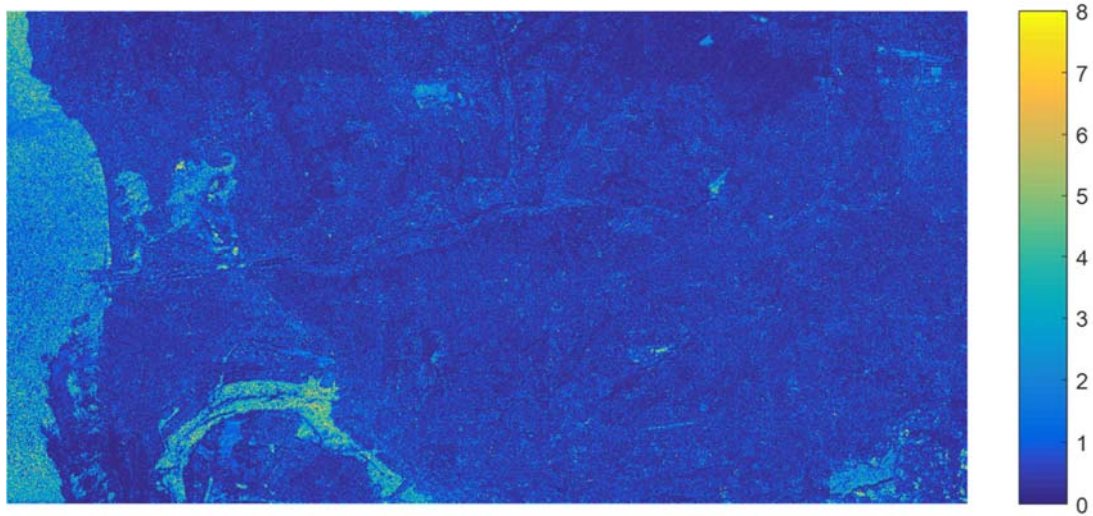

(a)

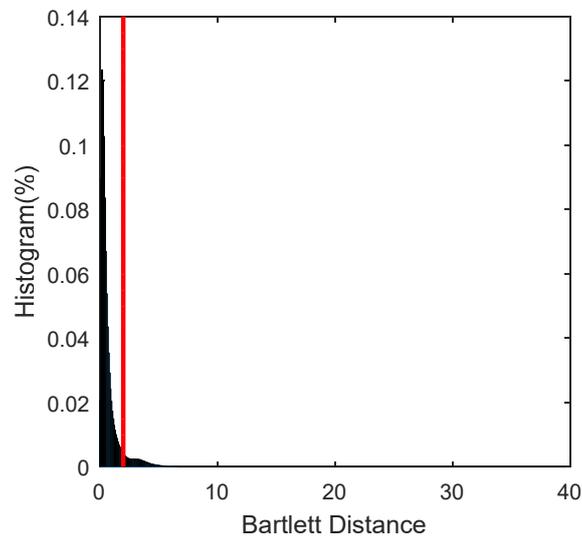

(b)

Figure 13 (a) Map of Bartlett distance between true and reconstructed full-pol SAR image of Test 1; (b) the corresponding histogram.

## c. *Applications*

The reconstructed full-pol data can now be used in the same way as true full-pol data in PolSAR applications. Polarimetric target decomposition is the most important PolSAR application. In this section, we show the how reliable of the reconstructed full-pol data if used in target decomposition. Model-based Freeman-Durden target decomposition [Freeman et al., 1998] is tested in our experiment.

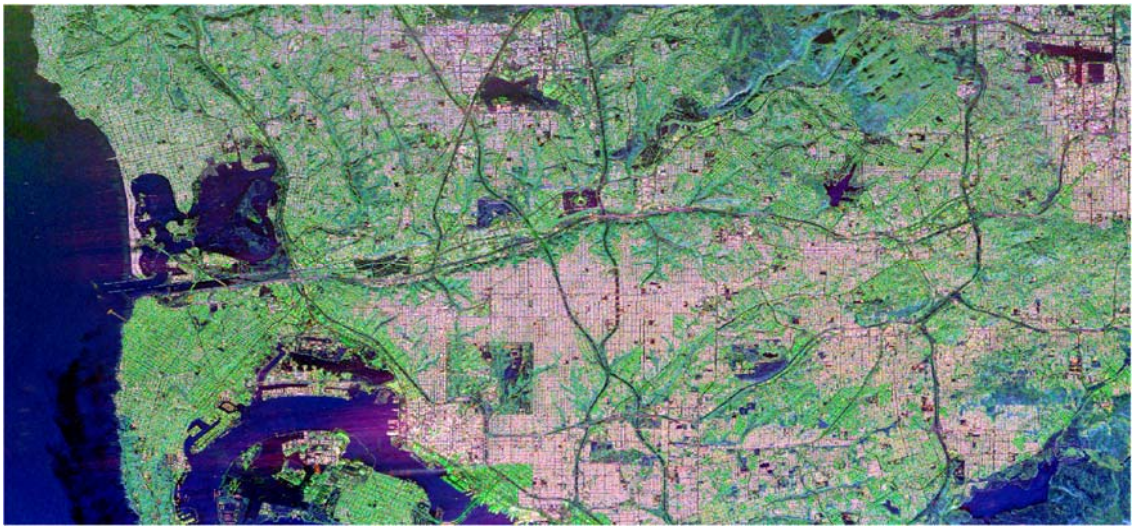

(a) Freeman-Durden decomposition of original true full-pol data.

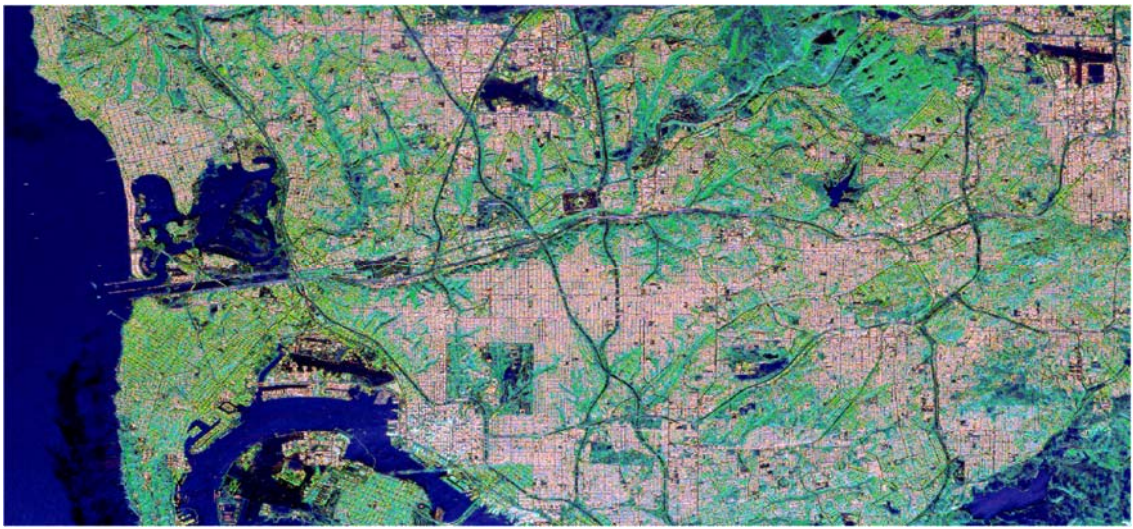

(a)  Freeman-Durden decomposition of reconstructed full-pol data

Figure 14 Freeman-Durden Decomposition

Fig. 14 (a) (b) shows the decomposition results of the original true full-pol data and the reconstructed full-pol data, respectively. The red, green and blue channel represents the double-bounce, volumetric and single-bounce scattering, respectively. Volumetric and double-bounce scatterers match almost perfectly. It's well-known that the volumetric component is easily overestimated in oriented built-up areas without deorientation [Xu et al., 2005]. This phenomenon is severe in Fig. 14(a), where some oriented urban blocks appears to be pure green. Interestingly, such phenomenon is slightly alleviated in the reconstructed image in Fig. 14(b), probably due to the fact that the reconstructed polarimetric feature is partially rooted from the spatial pattern, which is less sensitive to orientation than scattering characteristics. The surface component tends to be more homogeneous than that in Fig. 14(a). It suggests that reconstructed data is of good potential for unsupervised classification.

Eigen-analysis target decomposition proposed by Cloude and Pottier [Cloude and Pottier, 1997] is another commonly used PolSAR target decomposition. Fig. 15 compares the derived Cloude-Pottier decomposition parameter $H, \bar{\alpha}$, i.e. true value on the left vs. reconstructed value on the right on the one selected region. It seems that the derived polarimetric parameters agrees well between the reconstructed and true full-pol data, except some contamination areas. Meanwhile, the $H/\bar{\alpha}$ classification result of the reconstructed full-pol data agrees in most part with the true result. In fact, the classification result of the reconstructed full-pol seems to be more reasonable as it is not subjected to side-lobe contaminations in the cross-polarization channel.

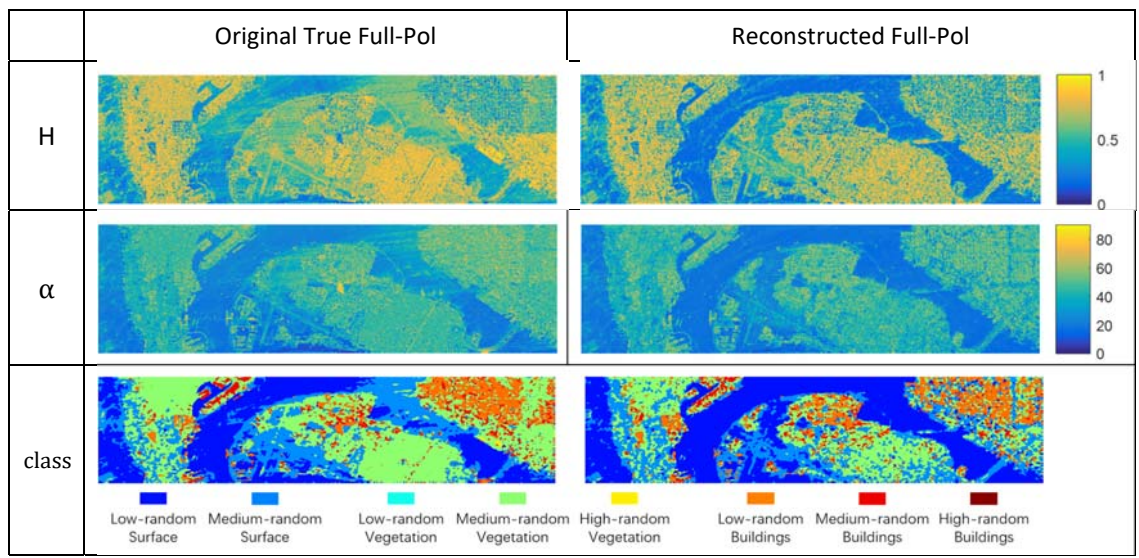

Figure 15 $H/\alpha$ unsupervised classification
25Submitted to IEEE TGRS

## 5. Conclusion

This paper proposes a novel DNN-based framework for converting single-pol SAR image to full-pol. This process is referred to as 'SAR image colorization'. It consists of two major DNNs, namely, the feature extractor network and the feature translator network. The former takes in the grayscale single-pol SAR image and extract the spatial features of different terrain types. The later translates spatial feature into polarimetric feature so that the covariance matrix of each pixel can be reconstructed. The feature extractor network can use layers from pretrained CNNs using optical images. However, the features have to be normalized specifically for SAR images. Since the feature extractor is fixed, the translator network has to be trained specifically for different SAR sensors and configuration/imaging modes. Note that the feature translator network employs softmax output layer which predicts the probability of 32 discrete levels of non-uniform quantization. This takes the advantage of superior performance of DNN as classifier but rather than regressor. Moreover, we also proposed an algorithm to correct for the semi-positive definiteness of the reconstructed covariance matrix. Visualization of the trained neural networks shows that the feature extractor network can effectively extract spatial features of different terrain types, and subsequently, the feature translator network can correctly map the spatial feature to the corresponding polarimetric feature.

Three full-pol UAVSAR images are used in the experiments, and three indicators are employed to quantitatively evaluate the performance of full-pol reconstruction. Moreover, the reconstructed full-pol data is also compared with original true full-pol data in terms of polarimetric target decompositions including both model-based and eigen-analysis-based decompositions. The results demonstrate that the reconstructed full-pol image can be used for both qualitative and quantitative PolSAR applications. Interestingly, we also found that the reconstructed polarization channel are free of contamination and interferences which could be potentially used for interference removal.

## References

R. Keith Raney, "Hybrid-Polarity SAR Architecture," IEEE Trans. On Geosci. and Remote Sens., vol. 45, no. 11, Nov. 2007.

A. Levin, D. Lischinski, and Y. Weiss, "Colorization Using Optimization," ACM SIGGRAPH, no. 689-694, 2004.

G. J. C. Souyris, P. Imbo, R. Fjørtoft, S. Mingot, and J.-S. Lee, "Compact polarimetry based on symmetry properties of geophysical media: The π/4 mode," IEEE Transaction on Geoscience and Remote Sensing, vol. 43, no. 3, pp. 634–646, March 2005.

T. Le, I. McLoughlin, K.Y. Lee and T. Bretschneider, "Neural Network-Assisted Reconstruction of Full Polarimetric SAR Information," Proceedings of the 4th International Symposium on Communications, Control and Signal Processing, March 2010.

K. Simonyan, A. Zisserman, "Very deep convolutional networks for large-scale image recognition," International Conference on Learning Representations, 2015.

G. Larsson, M. Maire, and G. Shakhnarovich, "Learning representations for automatic colorization," arXiv 1603.06668, 2016.

J. C. Souyris and S. Mingot, "Polarimetry based on one transmitting and two receiving polarizations: The pi/4 mode," IEEE International Geoscience and Remote Sensing Symposium, pp. 629–631, 2002.

O. Russakovsky, J. Deng, H. Su et al., "ImageNet Large Scale Visual Recognition Challenge," International Journal of Computer Vision, vol. 115, no. 3, 2015.

R. Zhang, P. Isola, and A. A. Efros, "Colorful Image Colorization," arXiv:1603.08511v2, 2016.

S. Chen et al., "Target Classification Using the Deep Convolutional Networks for SAR Images," IEEE Transaction on Geoscience and Remote Sensing, 2016.

S. A. Wagner, "SAR ATR by a combination of convolutional neural network and support vector machines," in IEEE Transactions on Aerospace and Electronic Systems, vol. 52, no. 6, pp. 2861-2872, December 2016.

Y. Zhou et al., "Polarimetric SAR Image Classification Using Deep Convolutional Neural Networks," IEEE Geoscience and Remote Sensing Letters, 2016.

L. Jiao and F. Liu, "Wishart Deep Stacking Network for Fast POLSAR Image Classification," in IEEE Transactions on Image Processing, vol. 25, no. 7, pp. 3273-3286, July 2016

Z. Zhang et al., "Complex-Valued Convolutional Neural Network and Its Application in Polarimetric SAR Image Classification," IEEE Transaction on Geoscience and Remote Sensing , 2017, In Press.

F. Xu, Q. Song and Y.Q. Jin, "Polarimetric SAR Image Factorization," IEEE Transaction on Geoscience and Remote Sensing, 2017, In Press.

F. Xu, Y.Q. Jin and A. Moreira, "A Preliminary Study on SAR Advanced Information Retrieval and Scene Reconstruction," IEEE Geoscience and Remote Sensing Letters, 2016.

F. Xu, Y. Li and Y.Q. Jin, "Polarimetric–Anisotropic Decomposition and Anisotropic Entropies of High-Resolution SAR Images," IEEE Transaction on Geoscience and Remote Sensing, 2016.

D.-X. Yue, F. Xu and Y. Q. Jin, "Wishart-Bayesian Reconstruction of Quad-Pol from Compact-Pol SAR Image," IEEE Geoscience and Remote Sensing Letters, Shanghai, 2016.

P. R. Kersten, J. S. Lee and T. L. Ainsworth, "Unsupervised Classification of Polarimetric Synthetic Aperture Radar Images Using Fuzzy Clustering and EM Clustering," IEEE Transactions on Geoscience and Remote Sensing, vol. 43, no. 3, Mar. 2005.

S. R. Cloude and E. Pottier, "An Entropy Based Classification Scheme for Land Applications of Polarimetric SAR," IEEE Transactions on Geoscience and Remote Sensing, vol. 35, no. 1, Jan., 1997.

A. Freeman and S. L. Durden, "A three-component scattering model for polarimetric SAR data," IEEE Transactions on Geoscience and Remote Sensing, vol. 36, no. 3, pp. 963–973, May 1998.

F. Xu and Y.-Q. Jin, "Deorientation theory of polarimetric scattering targets and application to terrain surface classification," IEEE Transactions on Geoscience and Remote Sensing, vol. 43, no. 10, pp. 2351–2364, Oct. 2005.

W. Liu, A. Rabinovich, A. C. Berg, "Parsenet: Looking wider to see better," arXiv: 1506.04579v2, Nov., 2015.

L.J.P. van der Maaten and G.E. Hinton, "Visualizing Data using t-SNE," Journal of Machine Learning Research, pp. 2579–2605, Nov., 2008.

C. D. Meyer, "Matrix Analysis and Applied Linear Algebra", SIAM, ISBN 0-89871-454-0.

Dataset: UAVSAR, NASA 2012. Retrieved from ASF DAAC Jan 2014.